# A PARALLEL IMPLEMENTATION OF THE COVARIANCE MATRIX ADAPTATION EVOLUTION STRATEGY

NAJEEB KHAN*

December 2015

**Abstract.** In many practical optimization problems, the derivatives of the functions to be optimized are unavailable or unreliable. Such optimization problems are solved using derivative-free optimization techniques. One of the state-of-the-art techniques for derivative-free optimization is the covariance matrix adaptation evolution strategy (CMA-ES) algorithm. However, the complexity of CMA-ES algorithm makes it undesirable for tasks where fast optimization is needed. To reduce the execution time of CMA-ES, a parallel implementation is proposed, and its performance is analyzed using the benchmark problems in PythOPT optimization environment.

**Key words.** optimization, parallel computation, evolutionary computation

**1. Introduction.** Optimization is the process of adjusting the inputs to or characteristics of a device, mathematical process, or experiment to find the minimum or maximum output or result. The input consists of variables; the process or function is known as the cost function, fitness function, or objective function; and the output is the cost or fitness [6].

For a point in the input space to be a local minimum (or a local maximum), the first-order necessary condition states that the derivative of the cost function must be zero. Thus, the derivative of a function includes important information for the optimization problem. Unfortunately, in many cases the derivatives of the objective functions are not available. For example, when the objective function is implemented using some commercial code for which only binary executable files are available or when the objective function is a large chunk of legacy code[1], extending such codes to get the first order derivatives is a prohibitively expensive task. Another example where derivatives are not available is the tuning of algorithmic parameters; many numerical codes for simulation, optimization, or estimation depend on a number of parameters [2]. Typically, these parameters are set to values that either have some mathematical justification or have been found by the code developers to perform well. This can be seen as an optimization problem with the objective function measuring the performance of the solver for a given choice of parameters. There are no derivatives available for such non-linear objective functions.

In this work, we consider black-box optimization problems, where the only information available about the objective function is its values (henceforth referred to as $f$-values) given some inputs. In such cases, the derivatives can be approximated by finite-differences, however, this is not feasible when the objective function evaluations are costly or when they are noisy.

To deal with such problems, derivative-free optimization (DFO) techniques are used [1]. Each DFO algorithm needs to perform a number of objective function evaluations, and based on the output values, it decides on a minimum value for the function. The best DFO algorithm will reach the minimum of the objective function with as few evaluations as possible, and thus for DFO algorithms, the number of objective function evaluations to reach a minimum point is used as a performance measure. One of the state-of-the-art DFO algorithms is the covariance matrix adaptation evolution strategy (CMA-ES) algorithm [5].

---

*najeeb.khan@usask.ca www.najeebk.com
[1]Code that was written in the past and is no longer maintained.
[2]For example, these could be the optimal number of processes for a specific algorithm to perform well.

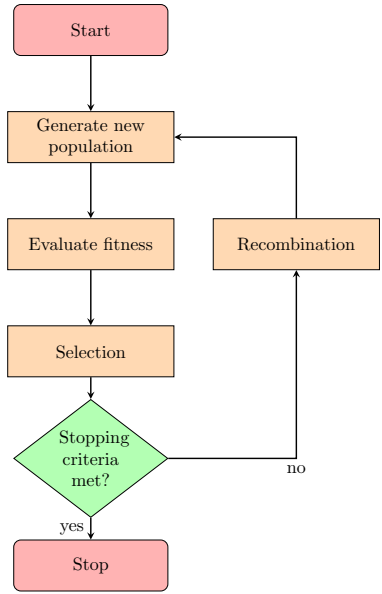

Fig. 1. *Flowchart of an Evolutionary algorithm.*

An interest in a parallel implementation of CMA-ES was aroused due to its use in the literature to optimize problems such as human gait modelling. In gait modelling, around 50 parameters are needed to be optimized for each frame. To simulate a 10 seconds gait at around 500 frames per second, around 5000 optimizations are required. If the CMA-ES is implemented in parallel, it will enable us to simulate human gait modelling for longer duration in less time.

**2. Covariance Matrix Adaptation Evolution Strategy.** Covariance Matrix Adaptation Evolution Strategy is a stochastic, derivative-free method for numerical optimization of non-linear or non-convex continuous optimization problems[3]. It belongs to the class of evolutionary algorithms and evolutionary computation.

An evolutionary algorithm is broadly based on the principle of biological evolution, namely the repeated interplay of variation (via recombination and mutation) and selection: as shown in Figure 1, in each generation (iteration) new individuals (candidate solutions, denoted as $\mathbf{x}$) are generated by variation, usually in a stochastic[4] way, of the current parental individuals. Then, some individuals are selected to become the parents in the next generation based on their objective function value $f(\mathbf{x})$. In this way, over the generation sequence, individuals with better and better $f$-values are generated. This iterative process is stopped when some stopping criteria is met such as the number of allowed objective function evaluations ($f$-evaluations) is reached or the desired fitness is achieved.

---

[3]Optimization problems where the objective function is non-linear and is defined over a continuous space rather than on some discrete points in the input space.

[4]The variation is not defined by a mathematical formula but a probability distribution.

In CMA-ES in each generation $K$ points $\mathbf{x}_1, \ldots, \mathbf{x}_K$ with dimension $n$ are sampled using a Gaussian probability distribution $\mathcal{N}(\mathbf{m}, \mathbf{C})$, where the Gaussian probability distribution is completely specified by the mean vector $\mathbf{m}$ and the covariance matrix $\mathbf{C}$. $Z$ of the $K$ points with the fittest $f$-values are selected and used to calculate a weighted mean for the Gaussian distribution of the next generation. The covariance matrix is calculated in such a way that both the variance between the selected points in one generation and the correlation between generations is fully exploited. The objective of CMA-ES is to fit the search distribution (the Gaussian probability distribution) to the contour lines[5] of the objective function to be minimized. As shown in the Figure 2 below, with each generation, the parameters of the search distribution, such as the mean vector and covariance matrix are adapted such that it coincides with the contour lines of the objective function and eventually all the $f$-evaluations are concentrated in the neighbourhood of the minimum.

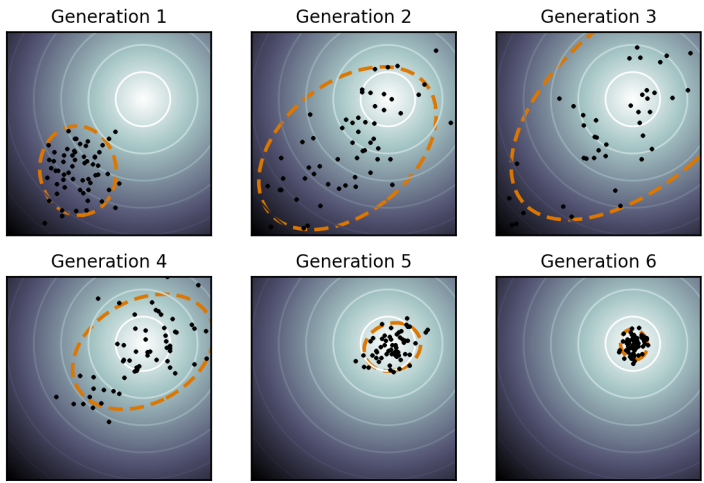

FIG. 2. *Fitting the search distribution to the contour lines of the objective function [11].*

**2.1. Sampling (Generating new population).** In each generation $g$, $K$ points $\mathbf{x}_1, \ldots, \mathbf{x}_K$ are sampled using a Gaussian probability distribution as follows

$$(2.1) \qquad \mathbf{x}_k^{(g+1)} \sim \mathbf{m}^{(g)} + \sigma^{(g)} \mathcal{N}(0, \mathbf{C}^{(g)}) \quad \text{for } k = 1, \ldots, K$$

where $\mathbf{m}^{(g)}$ and $(\sigma^{(g)})^2 \mathbf{C}^{(g)}$ are the mean vector and covariance matrix of the Gaussian distribution $\mathcal{N}$ at generation $g$, respectively. '$\sim$' denotes the same probability distribution on left and right. $\sigma^{(g)}$ is the step size and will be defined in later sections.

**2.2. Updating the mean (Selection and Recombination).** The objective function is evaluated at $K$ points generated in the previous step and $Z$ best points are selected. This corresponds to truncation selection in the evolutionary algorithm paradigm. A weighted average of the selected points is calculated as:

$$(2.2) \qquad \mathbf{m}^{(g+1)} = \sum_{i=1}^{Z} w_i \, \mathbf{x}_{i:K}$$

---

[5]Contours are points in the input space over which the value of the objective function is constant.

where $\mathbf{x}_{i:K}$ is the $i^{th}$ fittest point among $\mathbf{x}_1, \ldots, \mathbf{x}_K$ and $w_i$ are the recombination weights such that their sum equals one and $w_1 \geq w_2 \geq \cdots \geq w_Z > 0$. A parameter *variance effective mass* is defined as

$$(2.3) \qquad \mu_{\text{eff}} = \left( \sum_{i=1}^{Z} w_i^2 \right)^{-1}$$

If equal recombination weights are used, i.e., $w_i = 1/Z$, then $\mu_{\text{eff}} = Z$, more generally $1 \leq \mu_{\text{eff}} \leq Z$.

**2.3. Estimating the covariance matrix.** First we consider the estimation of the covariance matrix for the population of a single generation. The maximum likelihood estimate [2] of the covariance matrix given the Gaussian distributed samples $\mathbf{x}_1, \ldots, \mathbf{x}_K$ is given by

$$(2.4) \qquad \boldsymbol{C}_{\text{emp}}^{(g+1)} = \frac{1}{K-1} \sum_{i=1}^{K} \left( \mathbf{x}_i^{(g+1)} - \frac{1}{K} \sum_{j=1}^{K} \mathbf{x}_j^{(g+1)} \right) \left( \mathbf{x}_i^{(g+1)} - \frac{1}{K} \sum_{j=1}^{K} \mathbf{x}_j^{(g+1)} \right)^T$$

We can also use the original mean $\mathbf{x}^{(g)}$ which generated the samples instead of the sample mean in 2.4,

$$(2.5) \qquad \boldsymbol{C}_K^{(g+1)} = \frac{1}{K} \sum_{i=1}^{K} \left( \mathbf{x}_i^{(g+1)} - \mathbf{m}^{(g)} \right) \left( \mathbf{x}_i^{(g+1)} - \mathbf{m}^{(g)} \right)^T.$$

The key difference between 2.4 and 2.5 is the reference mean value. For $\boldsymbol{C}_{\text{emp}}^{(g+1)}$ we use the mean value of the realized sample $\frac{1}{K} \sum_{j=1}^{K} \mathbf{x}_j^{(g+1)}$, while for $\boldsymbol{C}_K^{(g+1)}$ we use the true mean $\mathbf{m}^{(g)}$ of the sampled distribution. $\boldsymbol{C}_{\text{emp}}^{(g+1)}$ can be interpreted as an estimate of the variance within the sampled points, while $\boldsymbol{C}_K^{(g+1)}$ estimates the variances of the *sampled steps* $\mathbf{x}_i^{(g+1)} - \mathbf{m}^{(g)}$. 2.5, however, gives us a re-estimate of the original covariance matrix $\boldsymbol{C}^{(g)}$, we use the same recombination procedure used for the mean update in 2.2, to get a better estimate in terms of objective function fitness.

$$(2.6) \qquad \boldsymbol{C}_Z^{(g+1)} = \sum_{i=1}^{Z} w_i \left( \mathbf{x}_{i:K}^{(g+1)} - \mathbf{m}^{(g)} \right) \left( \mathbf{x}_{i:K}^{(g+1)} - \mathbf{m}^{(g)} \right)^T.$$

Just as the variance in the sampled steps is estimated by $\boldsymbol{C}_K^{(g+1)}$, the variance of the *selected steps* is estimated by $\boldsymbol{C}_Z^{(g+1)}$. In 2.6, the covariance matrix is estimated using a single population. When the population size is small, 2.6 results in unreliable estimates. To deal with this problem, information from previous generations can be used; for example, the next generation covariance matrix can be estimated as the average of all the covariance matrices of the previous generations.

$$(2.7) \qquad \boldsymbol{C}_Z^{(g+1)} = \frac{1}{g+1} \sum_{i=0}^{g} \frac{1}{\sigma^{(i)2}} \boldsymbol{C}_Z^{(i+1)}$$

A better approach would be to assign recent generations a higher weight rather than equal weight for all the generations. To accomplish this, exponential smoothing[6]

---

[6]Exponential smoothing is an averaging technique which assigns exponentially decreasing weights as the observations get older and is given by $y_t = cx_t + (1-c)y_{t-1}$.

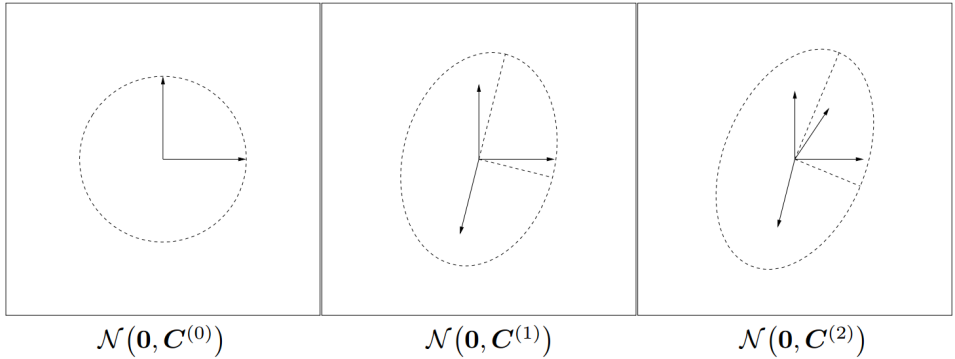

Fig. 3. *Change of the distribution according to the covariance matrix update 2.10 [4]*

with a learning rate $0 < c_Z \leq 1$ is used, and the resulting equation is known as rank-$Z$-update

$$(2.8) \qquad \boldsymbol{C}^{(g+1)} = (1-c_Z)\boldsymbol{C}^{(g)} + c_Z \frac{1}{\sigma^{(g)2}} \boldsymbol{C}_Z^{(g+1)}$$

$$(2.9) \qquad \boldsymbol{C}^{(g+1)} = (1-c_Z)\boldsymbol{C}^{(g)} + c_Z \sum_{i=1}^{Z} w_i \boldsymbol{y}_{i:K}^{(g+1)} \boldsymbol{y}_{i:K}^{(g+1)T}$$

where $\boldsymbol{y}_{i:K}^{(g+1)} = \left(\boldsymbol{x}_{i:K}^{(g+1)} - \boldsymbol{m}^{(g)}\right)/\sigma^{(g)}$.

Apart from the rank-$Z$-update, CMA-ES also uses a rank-1-update in which the covariance matrix is repeatedly updated in the generation sequence using a single *selected step* only. Setting $Z$ in 2.9 equal to 1, we get

$$(2.10) \qquad \boldsymbol{C}^{(g+1)} = (1-c_1)\boldsymbol{C}^{(g)} + c_1 \boldsymbol{y}^{(g+1)} \boldsymbol{y}^{(g+1)T}.$$

The outer product on the right of the summand is of rank one and adds the maximum likelihood term for $\boldsymbol{y}^{(g+1)}$ into the covariance matrix $\boldsymbol{C}^{(g)}$, thus increasing the probability of generating $\boldsymbol{y}^{(g+1)}$ in the next generation. Figure 3 shows the first two iterations of 2.9 with $c_1 = 0.17$. In the first iteration $\boldsymbol{C}^{(0)}$ is set to the identity matrix $\boldsymbol{I}$. In the middle of Figure 3, $\boldsymbol{y}^{(1)} = [-0.59, -2.2]^T$ is the selected step and the axis of $\boldsymbol{C}^{(1)}$ can be seen to align to the axis of the selected step, in the right of Figure 3, $\boldsymbol{y}^{(2)} = [0.97, 1.5]^T$ is the selected step and the axis of $\boldsymbol{C}^{(2)}$ further rotates clockwise to align to the selected step.

One thing to note about 2.10 is that the sign of the selected steps does not count, as $\boldsymbol{yy}^T = (-\boldsymbol{y})(-\boldsymbol{y})^T$ and thus to utilize the sign information, the concept of an evolution path is introduced. An evolution path is a sequence of successive steps, that the strategy takes over a number of generations. An evolution path can be expressed by a sum of consecutive steps. This summation is referred to as cumulation [4]. An evolution path of three steps of the distribution mean $\boldsymbol{m}$ can be constructed by the sum

$$(2.11) \qquad \frac{\boldsymbol{m}^{(g+1)} - \boldsymbol{m}^{(g)}}{\sigma^{(g)}} + \frac{\boldsymbol{m}^{(g)} - \boldsymbol{m}^{(g-1)}}{\sigma^{(g-1)}} + \frac{\boldsymbol{m}^{(g-1)} - \boldsymbol{m}^{(g-2)}}{\sigma^{(g-2)}}$$

A better strategy would be to use exponential smoothing similar to 2.10 to get the evolution path $\boldsymbol{p}_c^{(g)}$ given by

$$(2.12) \qquad \boldsymbol{p}_c^{(g+1)} = (1-c_c)\boldsymbol{p}_c^{(g)} + \sqrt{c_c(2-c_c)\mu_{\text{eff}}}\,\frac{\mathbf{m}^{(g+1)} - \mathbf{m}^{(g)}}{\sigma^{(g)}},$$

where $c_c$ is the backward time horizon for the evolution path. The normalization constant $\sqrt{c_c(2-c_c)\mu_{\text{eff}}}$ ensures that the $\boldsymbol{p}_c^{(g+1)}$ and $\boldsymbol{p}_c^{(g)}$ has the same probability distribution.

Combining 2.9, 2.10 and 2.12 we finally arrive at the CMA update equation

$$(2.13) \qquad \boldsymbol{C}^{(g+1)} = (1-c_1-c_Z)\,\boldsymbol{C}^{(g)} + c_1\underbrace{\boldsymbol{p}_c^{(g+1)}\boldsymbol{p}_c^{(g+1)^T}}_{\text{rank one update}} + c_Z\underbrace{\sum_{i=1}^{Z} w_i\,\mathbf{y}_{i:K}^{(g+1)}\mathbf{y}_{i:K}^{(g+1)^T}}_{\text{rank}-Z-\text{update}}$$

In practice $c_1 \approx 2/n^2$ and $c_Z \approx min(\mu_{\text{eff}}/n^2, 1-c_1)$ [4].

**2.4. Step size control.** The only task remaining now is that of the step size $\sigma^{(g+1)}$ selection. Consider Figure 4, where three different selection scenarios are shown. Although the length of the individual steps are the same, the length of the evolution path is quite varied. In the left side, the evolution path is short, loosely speaking, the individual steps cancel out each other i.e. they are anti-correlated and to remedy the situation, the step size should be decreased. On the right side of Figure 4, the evolution path is long, the individual steps are in almost the same direction i.e. they are correlated and a larger step size can be used with fewer steps to cover the same distance. The middle of Figure 4 shows the desired length of the evolution path where the individual steps are approximately uncorrelated and hence there is no need to change the step size in this case.

Now the question is how do we decide whether an evolution path is long or short? To deal with this, we compare the length of the evolution path with the expected length of the path under random selection, in which the consecutive steps are independent[7] and thus uncorrelated[8] and this is what the preferred path length should be. Thus we increase the step size if the evolution path is longer than the random selection path and vice versa.

The evolution path defined by 2.12 depends on the direction of the individual steps, however, we need to find an evolution path independent of the direction, to deal with this a conjugate path $\boldsymbol{p}_c^{(g+1)}$ which is independent of the direction of the selected steps is constructed as

$$(2.14) \qquad \boldsymbol{p}_\sigma^{(g+1)} = (1-c_\sigma)\boldsymbol{p}_\sigma^{(g)} + \sqrt{c_\sigma(2-c_\sigma)\mu_{\text{eff}}}\,\boldsymbol{C}^{(g+1)^{-\frac{1}{2}}}\,\frac{\mathbf{m}^{(g+1)} - \mathbf{m}^{(g)}}{\sigma^{(g)}},$$

where the factor $\boldsymbol{C}^{(g+1)^{-\frac{1}{2}}}$ acts as a whitening transform[9], making the expected length of $\boldsymbol{p}_\sigma^{(g+1)}$ independent of its direction. The step size is then selected based on the comparison of the length of the conjugate path $\boldsymbol{p}_\sigma^{(g+1)}$ with the expected length of a Gaussian variable with zero mean and identity covariance matrix $\mathcal{N}(\boldsymbol{0}, \mathbf{I})$ as follows.

---

[7] Two random variables $X$ and $Y$ are independent when their joint probability distribution is the product of their marginal probability distributions i.e. $p_{X,Y}(x,y) = p_X(x)p_Y(y)$.

[8] If $p_{X,Y}(x,y) = p_X(x)p_Y(y)$ then the joint expectation is $E[X,Y] = E[X]E[Y]$ and by definition the covariance $cov(X,Y) = E[X,Y] - E[X]E[Y]$ is zero.

[9] Whitening transforms an arbitrary set of variables having a known covariance matrix **C** into a set of new variables whose covariance is the identity matrix meaning that they are uncorrelated and all have variance 1, a brief mathematical explanation can be found in [10].

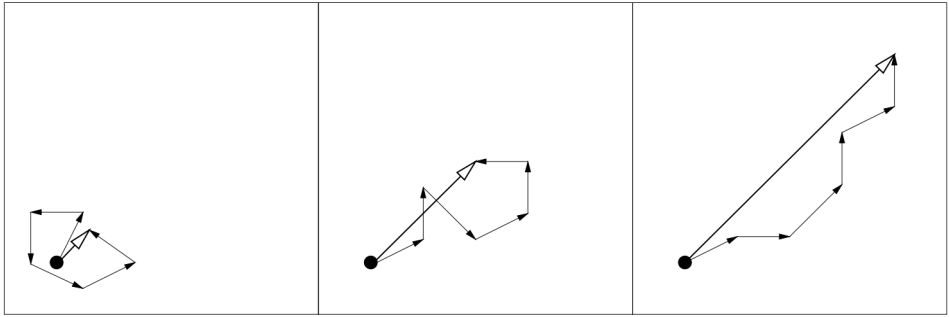

FIG. 4. *Three evolution paths of six (equal length) steps from different selection situations [4]*

$$(2.15) \qquad \sigma^{(g+1)} = \sigma^{(g)} \exp\left(\frac{c_\sigma}{d_\sigma}\left(\frac{\|\boldsymbol{p}_\sigma^{(g+1)}\|}{E\|\mathcal{N}(0,I)\|} - 1\right)\right)$$

where $d_\sigma \approx 1$ is a damping coefficient and scales the changes in of step size magnitude.

**2.5. The Algorithm.** Pseudo-code of the CMA-ES algorithm is given in Algorithm 1.

---
**Algorithm 1** CMA-ES
---
    Set algorithmic parameters to default values given in Table 1 of [4]
    Set evolution paths $\boldsymbol{p}_\sigma = \boldsymbol{0}$, $\boldsymbol{p}_c = \boldsymbol{0}$, covariance matrix $\boldsymbol{C} = \boldsymbol{I}$, and $g = 0$
    Initialize mean $m \in \mathbb{R}^n$, step size $\sigma$
    **while** [NOT DONE] **do**
      **for** $i \Leftarrow 1$ to $K$ **do**
        $\mathbf{x}_i \Leftarrow \mathcal{N}(\mathbf{m}, \mathbf{C})$ $\{\mathbf{x}_i \in \mathbb{R}^n\}$
        $f_i \Leftarrow Objective(\mathbf{x}_i)$
      **end for**
      $Sort(\mathbf{f}, \mathbf{x})$
      $\boldsymbol{m} \Leftarrow mean(\mathbf{x}_1 : \mathbf{x}_Z)$ using 2.2
      Update $\boldsymbol{p}_\sigma$ using 2.14
      Update $\boldsymbol{p}_c$ using 2.12
      Update $\boldsymbol{C}$ using 2.13
      Update $\sigma$ using 2.15
      Determine if DONE?
    **end while**
    Return $\mathbf{m}, Objective(\mathbf{m})$
---

**3. Serial Implementation.** In order to implement a parallel CMA-ES we take the two step approach; we implement CMA-ES sequentially in the pythOpt optimization environment and then parallelize the sequential implementation.

**3.1. The pythOpt Optimization Framework.** The pythOpt [12] is an optimization framework that simplifies the use of certain optimization packages such as VTDirect95 and LGO. Besides, native python implementations of the variants of the particle swarm optimization (PSO) algorithm, pythOpt also includes several optimization problems and benchmark functions, which can be used to evaluate different optimization solvers. The pythOpt provides a unified interface for implementation of new solvers and objective functions both in the python and the FORTRAN programming languages.

In order to include a new python solver in pythOpt a new class inheriting the `PythonSolver` class need to be included. The new solver class should, at the least, provide implementation for the `solve()` function as shown in Listing 1.

LISTING 1. *Minimal code for adding a new optimization solver to pythOpt*
```python
from ..engine.solver    import PythonSolver
from ..engine.solution import Solution
class NewSolver(PythonSolver):
    """ DocString of the new solver here
    """
    DETERMINISTIC = False # deterministic vs stochastic?

    def __init__(self):
        "Constructor for the new solver class \
         Initialize the parameters of the algorithm here."
        PythonSolver.__init__(self)

    def solve(self):
        "Implementation of the new solver here."
        self.solution = Solution({minimum:-123, position:[0,0]})
        self._solved = True
        return self.solution
```

A 'well-commented', vectorized implementation of the CMA-ES algorithm using the pythOpt framework and the numpy library is shown in Listing 5 at the end of this document.

**3.2. Example Use and Functional Correctness of the Serial Implementation.** In order to solve an optimization problem in pythOpt, first define an objective function that takes the sample point $x$ in the search space and returns a dictionary containing the $f$-value and the search point $x$[10]. Listing 2 shows the sphere objective function $f(x) = \sum_i x_i^2$.

LISTING 2. *The sphere objective function in python*
```python
def sphere(xs):            #objective function
    res = sum(map(lambda x: x*x, xs))
    sol = dict(minimum=res, position=list(xs))
    return sol
```

---

[10]This may be to keep the correspondence between points and their fitness in non-deterministic parallel evaluation?

The code to find the minimum point of the $N$-dimensional sphere function is shown in Listing 3 and the output of the code is shown in Figure 5.

LISTING 3. *pythOpt code for optimization using different algorithms*
```python
from pythopt import get_solver, CubeSpace, Objective
N=10 #problem dimension
settings = {'eval_lim' : 500, #max number of function evaluations
            'seed' : 123  #seed value for random number generator
}
bounds=(-20, 20) #search space bounds
obj    = Objective(sphere) #Get a pythOpt wrapper for objective function
space  = CubeSpace(N, bounds) #Get the search space
solution={}
# testing different solvers
for solver_name in ['PSO', 'DWPSO', 'TVACPSO', 'CMAES']:
        solver = get_solver(solver_name)
        solver_ready = solver(obj, space,settings=settings)
        solution[solver_name] = solver_ready.solve()
        print solver_name + ' : '+str(solution[solver_name]['minimum'])
```

```
PSO : 43.5430253063
DWPSO : 11.161322423
TVACPSO : 12.4502601472
CMAES : 0.0443865847057
```

FIG. 5. *Output of the code in Listing 3*

For the simple sphere objective function, as can be seen in Figure 5, the CMA-ES implementation outperforms all the PSO algorithms. This evidence to some extent supports the functional correctness of the CMA-ES implementation. To further test the serial implementation, standard global optimization benchmark functions were used. Results of the benchmark problems are given in Tables 1 and 2. Details of the benchmark functions can be found in [7]. As can be seen in Table 1, given a fixed number of $f$-evaluations, CMA-ES performs better than all the variants of the Particle Swarm Optimization (PSO) algorithms, in most of the cases[11]. Among the PSO variants, the Time Varying Acceleration Coefficients Particle Swarm Optimization (TVACPSO) performs better, so CMA-ES is compared with TVACPSO for different number of $f$-evaluations in Table 2.

---

[11]I am not claiming that CMA-ES is better than the other algorithms, I just want to indicate that the implementation is not too buggy and produces comparable or better results.

TABLE 1
*Comparison of the optimization algorithms: function evaluations limit set to 1024*

|  | CMA-ES | PSO | DWPSO | TVACPSO |
|---|---:|---:|---:|---:|
| easom | 0 | 0.0059648 | 1.02274e-09 | 1.33868e-08 |
| dejongfive | 16.3764 | 3.96004 | 3.95249 | 0.994027 |
| sphere | 1580.27 | 17277.9 | 18672.2 | 21431.3 |
| shubert | 1.98952e-13 | 0.00803123 | 22.5725 | 1.44454e-07 |
| noncontinousrastrigin | 348.367 | 155.908 | 198.051 | 169.077 |
| quadricnoise | 0.994001 | 0.558324 | 0.665178 | 1.30158 |
| goldsteinprice | 1.77636e-14 | 0.00830766 | 2.63141e-08 | 4.02083e-11 |
| deceptive | -0 | -0 | -0 | -0 |
| hyperellipsoid | 0 | 0 | 0 | 0 |
| camelback | 2.22045e-16 | 0.00109194 | 3.86567e-09 | 5.16458e-10 |
| michalewicz | 7.56838 | 3.61357 | 3.31855 | 4.11536 |
| rastigrin | 348.985 | 241.288 | 199.433 | 223.584 |
| parabola | 1893.56 | 1464.57 | 1266.13 | 1135.44 |
| schwefelp2_22 | 4.60231 | 31.0313 | 30.1729 | 27.7766 |
| tripod | 4.03589e-13 | 1.72912 | 1.00103 | 0.000356884 |
| step | 5 | 4363 | 3852 | 3886 |
| griewank | 0.692574 | 12.0602 | 9.3359 | 8.38043 |
| ackley | 1.73407 | 5.75319 | 5.38469 | 4.71177 |
| schaffersf6 | 0.0781892 | 0.0123266 | 0.00971592 | 0.00971591 |
| alpine | 0.0383044 | 3.77436 | 1.86649 | 2.0443 |
| rosenbrock | 111.772 | 6705.22 | 4468.04 | 2536.83 |
| generalizedpenalized | 0.326737 | 23592.4 | 40.7071 | 152.227 |
| dropwave | 0.0637547 | 0.0638314 | 0.0637547 | 0.0637547 |

**4. Parallel Implementation.**

**4.1. Related Work.** In [9], the authors combines the PSO and CMA-ES algorithms, considering multiple instances of the CMA-ES as particles in the PSO algorithm, and parallelize the CMA-ES instances with different initial conditions. A rough sketch of this approach is shown in Figure 6. In [3], instead of using multiple instances, the authors have proposed a parallel implementation of the objective function evaluations in a signal CMA-ES instance. This is shown in Figure 7.

**4.2. Parallelization.** It is assumed that the $f$-evaluations are expensive compared to the dimension of the problem and hence to the complexity of CMA-ES update. An approach similar to that of [3] is used to parallelize the $f$-evaluations in a single instance of the CMA-ES.

Parallel programming in Python is supported by the `mpi4py` and `multiprocessing` packages. Since the parallelization needed to be in the pythOpt framework, a fork join paradigm was more favorable to the single program multiple data paradigm used by `mpi4py`. However, `multiprocessing` support only functions that can be pickled[12]. Unfortunately, pythOpt uses a very nested structure and the objective function can not be pickled. To deal with this problem, `pathos`, a library that provides the communication mechanisms for configuring and launching parallel computations across heterogenous resources[8], was used. FORTRAN objective functions are still not pickable and thus cannot be used with our parallel implementation. The parallel CMA-ES code is shown in Listing 6.

---

[12]Serialiazation or marshalling is the process of transforming the memory representation of an object to a data format suitable for storage or transmission, and it is typically used when data must be moved from one program to another. Serialization in python is supported by the 'pickle' module and is known as pickling.

TABLE 2
*Difference between the minimum obtained by TVACPSO and CMA-ES. The values in red shows TVACPSO obtaining a better minimum than CMA-ES.*

| Number of $f$-evals | 512 | 1024 | 5000 | 10000 |
|---|---|---|---|---|
| easom | 1.92703e-05 | 1.33868e-08 | 0 | 0 |
| dejongfive | -15.1438 | -15.3824 | -10.6785 | -10.6785 |
| sphere | 11809.8 | 19851 | 16650.4 | 17888.1 |
| shubert | 0.042657 | 1.44453e-07 | -2.84217e-14 | -132.326 |
| noncontinousrastrigin | -182.63 | -179.29 | 24.9296 | 94.526 |
| quadricnoise | -10.4552 | 0.307576 | 0.00448389 | -0.124305 |
| goldsteinprice | 2.78632e-05 | 4.01905e-11 | -1.08802e-13 | -4.9738e-14 |
| deceptive | 0 | 0 | 0 | 0 |
| hyperellipsoid | 0 | 0 | 0 | 0 |
| camelback | 2.8832e-06 | 5.16457e-10 | 0 | 0 |
| michalewicz | -1.668 | -3.45302 | 1.16768 | 0.247294 |
| rastigrin | -131.71 | -125.4 | 190.303 | 166.327 |
| parabola | -1435.17 | -758.123 | 1256.12 | 1261.39 |
| schwefelp2_22 | 0.6144 | 23.1743 | 22.8491 | 25.3348 |
| tripod | 0.0351779 | 0.000356884 | 1 | 1.15276e-42 |
| step | 4202 | 3881 | 2244 | 2587 |
| griewank | 8.16214 | 7.68786 | 5.48973 | 6.5326 |
| ackley | 2.52154 | 2.9777 | 5.07406 | 5.21333 |
| schaffersf6 | 0.0275123 | -0.0684733 | -0.126991 | -0.0684733 |
| alpine | 0.805608 | 2.006 | 2.10865 | 1.94159 |
| rosenbrock | 16487.7 | 2425.06 | 5752.57 | 11075.3 |
| generalizedpenalized | 72.8681 | 151.9 | 30.3723 | 244.28 |
| dropwave | -0.317727 | 2.8307e-11 | -0.0637547 | -0.0637547 |

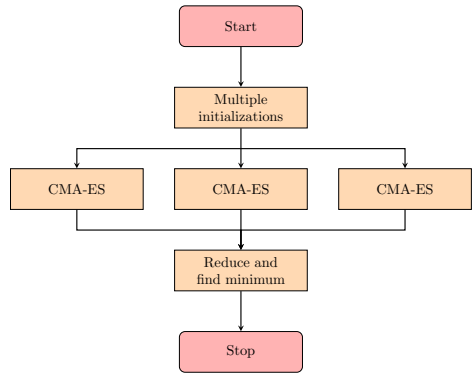

FIG. 6. *Parallelization of CMA-ES in [9]*

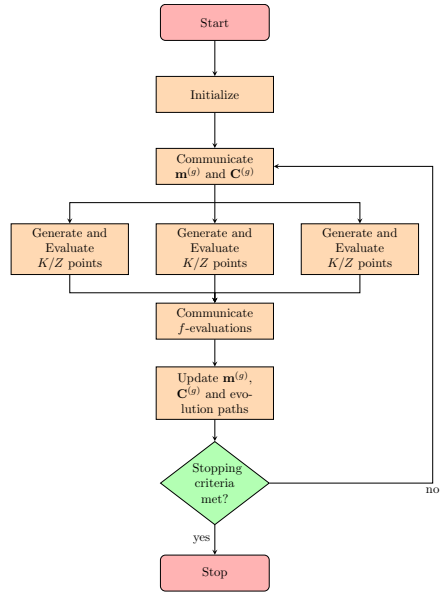

FIG. 7. *Parallelization of CMA-ES in [3]*

LISTING 4. *An expensive objective funtion with dummy computation*
```
def dummySphere(xs):   # dummy expensive objective function
    res = sum(map(lambda x: x*x, xs))
    sol = dict(minimum=res, position=list(xs))
    s=0
    for i in range(1,int(1e8)): #dummy summation
        s += 1.0/i
    return sol
```

**4.3. Results.** To test the performance of the parallel CMA-ES, due to lack of time, dummy computation was added to the sphere objective function as shown in Listing 4. Description of the parameters used in the experiments is shown in Table 3. Execution times for different number of processors is shown in Figure 8, while speedup and efficiency are shown in Figure 9. Since the objective function is still very simple, the efficiency is quite poor; however, if a real world expensive objective function is used, the efficiency will improve.

**5. Future Work.** In literature, eigen-decomposition is used to compute the inverse square root of the covariance matrix. Algorithms for Eigen-decomposition generally have a complexity of $O(N^3)$. Since the covariance matrix is symmetric positive semi-definite (PSD), we can use Cholesky decomposition to compute a triangular square root matrix and then invert it, A complexity analysis of the existing and the Cholesky decomposition method may reveal advantages of Cholesky decomposition in terms of parallelization and scalability. This idea is shown in Figure 10.

**6. Acknowledgement.** Thanks to Dr. Raymond Spiteri for the thorough and useful feedback on this report. My understanding of parallelism is due to him, credit him for everything that is correct, blame me for all the inaccuracies.

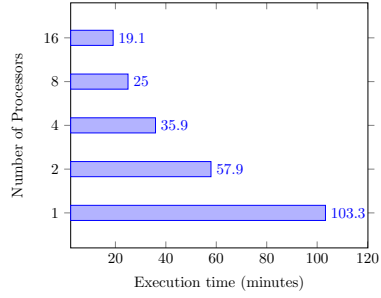

Fig. 8. *Execution times of PCMA-ES.*

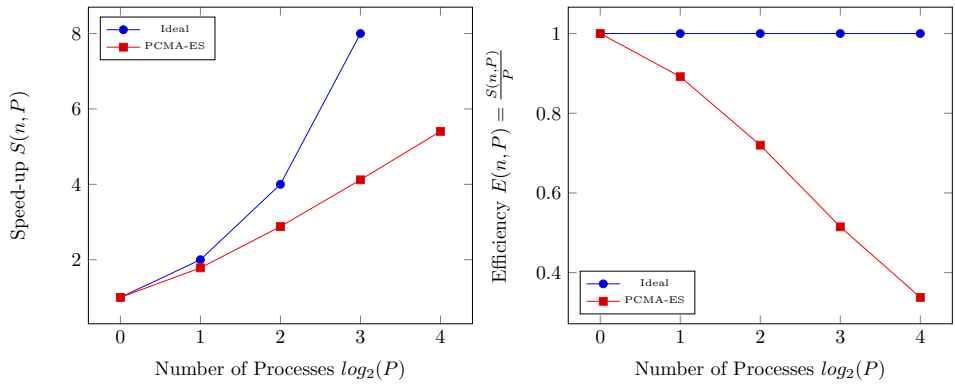

Fig. 9. *Speedup and efficiency of the PCMA-ES defined as $S(n.P) = \frac{T(n,1)}{T(n,P)}$ and $E(n.P) = \frac{S(n,P)}{P}$, respectively.*

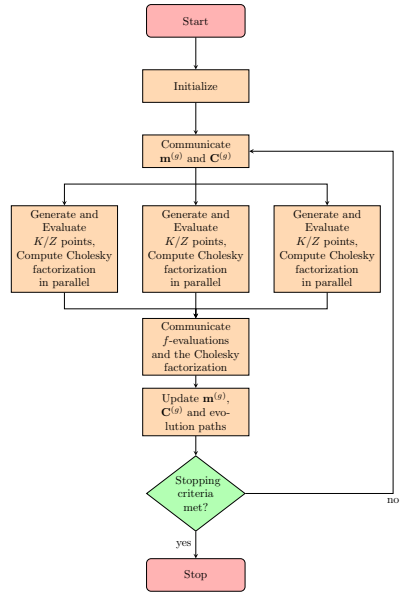

Fig. 10. *Finer parallelization of the CMA-ES*

TABLE 3
*Parameters used for the performance analysis of the parallel implementation*

| Variable | Value |
|---|---|
| Maximum number of $f$-evals | 512 |
| Problem dimension $n$ | 100 |
| Single $f$-eval time | 13.05 |

LISTING 5
*Serial implementation of CMA-ES*

```python
"""Covariance Matrix Adaptation Evolution Strategy

Implementation of CMAES algorithm based on

[1] Hansen, N. (2005). The CMA evolution strategy: A tutorial. Vu le, 29.
[2] Khan, N. (2015). A parallel implementation of the CMA-ES. Unpublished

.. moduleauthor:: Najeeb Khan <najeeb3.141@gmail.com>
"""
from ..engine.objective import FortranObjective
from ..engine.solution import Solution
from ..engine.solver    import PythonSolver
from ..engine.config    import Config
from ..engine.problemdata import ProblemData
from ..engine.checkpoint import checkpoint
import numpy as np
import scipy.linalg as la
import unicodedata
import logging
LOGGER = logging.getLogger(__name__)

class CMAES(PythonSolver):
  """ Covariance Matrix Adaptation Evolution Strategy
  """
  settings = {
    'xInit'    : np.array([]), # Initial guess
    'sigma'    : 1,            # Initial step size
    'seed'     : 0,            # Default seed value
    'eval_lim' : 100           # Stopping Criteria
  }

  DETERMINISTIC = False        # CMA-ES is stochastic

  def __init__(self, obj, space, settings=dict()): # constructor
    PythonSolver.__init__(self)
    np.random.seed(settings['seed'])
    # CMA-ES doesn't need this space as the mean and covariance matrix
    # completely specify the Gaussian PDF

    if space:                       # We use it for getting the problem dimenstion,
      bounds=space.get_bounds()     # initial guess xInit, and the step-size sigma.
      self.N=len(bounds)
      bounds=bounds[0]
```

```python
    # From Intuition (The best I could think of given the time!)
    # this will be overridden by user supplied initial step size
    #                      -----------------------------
    #initial step size = \|(UpperLimit - LowerLimit)*0.5
    #
    self.settings['sigma']=np.sqrt((bounds[1]-bounds[0])/2.0)

    #Initial guess xInit
    # xInit = mean point of the space + step size * N(0,I)
    self.settings['xInit']=((bounds[1]+bounds[0])/2.0) * np.ones(self.N) + \
      np.random.multivariate_normal(np.zeros(self.N),np.eye(self.N))

#Update settings
self.settings = dict(self.settings)
self.settings.update(settings)
self.objFunc = obj
self.N=len(self.settings['xInit'])
self.mean=np.copy(self.settings['xInit'])
self.sigma=self.settings['sigma']

# Population size set according to Eq.44 in [1]
self.popSize=4 + int(3 * np.log(self.N))

# Half of the population is selected for recombination [1]
self.mu=self.popSize/2

# Set the weights W according to Eq.45 in [1]
self.W=np.arange(1,self.mu+1)
self.W=np.log(self.mu+0.5) - np.log(self.W)
self.W=self.W/np.sum(self.W)
self.muEff = 1 / np.dot(self.W, self.W)

# Evolution path smoothing constant set according to Eq.47 in [1]
self.cCumm = (4 + self.muEff/self.N) / (self.N + 4 + 2*self.muEff/self.N)
# Direction independent evolution path smoothing constant
# set according to Eq.46 in [1]
self.cSig = (self.muEff + 2) / (self.N + self.muEff + 5)
# Rank-1-update smoothing constant set according to Eq.48 in [1]
self.cOne = 2 / (np.square(self.N + 1.3) + self.muEff)
# Rank-mu_update smoothing constant Eq.49 in [1]
self.cMu = np.min([ 1-self.cOne , 2*(self.muEff - 2 + 1/self.muEff) / \
    (np.square(self.N + 2) + self.muEff)] )
# damping factor for step size update set according to Eq.46 in [1]
self.dSig = 2.0 * self.muEff/self.popSize + 0.3 + self.cSig
# Evolution path vector
self.evolPath = np.zeros(self.N)
# Direction independent evolution path vector
self.sigPath = np.zeros(self.N)
# Matrix of eigenvectors of the covariance matric C
self.B = np.eye(self.N)
# Vector of eigenvalues of the covariance matrix C
self.D = np.ones(self.N)
# The covariance matrix
```

```python
    self.C = np.eye(self.N)
    # The inverse square root of the covariance matrix
    self.invSqrtC = np.eye(self.N)
    # counting variables
    self.eigenEval = 0
    self.countEval = 0
    self.gNum=0
    self.fitvals = []
    # To make pythOpt happy
    self.result=()
    self._solved = False

def solve(self):
  # while stopping criteria not met
  while self.countEval <= self.settings['eval_lim']:
    X, fVal = self.generatePop()              # Generate new population
    self.updateCMA(X, fVal)                   # Update CMA
  self.solution = Solution(self.result)       # respect pythOpt
  self._solved = True                         # Declare victory
  return self.solution                        # Pass the solution!

def generatePop(self):
  """
  Function to generate new candidate solutions according to the Gaussian
  PDF, evaluate fitness on these points and also decompose the covariance
  in order to find its inverse square root.
  """
  # Increment generation number
  self.gNum += 1

  # Decompose the covariance matrix into BDB'
  try:
    self.D,self.B = la.eigh(self.C)
    self.eigenEval = self.countEval
  except Exception, e:
    print "Covariance matrix bad beyond machine precisoin!"
    sys.exit(1)

  # Compute the inverse square root as B' D^-0.5 B
  self.D=np.sqrt(self.D)
  tmpD=np.diag(1.0/(self.D + np.finfo(float).eps)) # I hope eps wouldn't hurt
  self.invSqrtC = np.dot(self.B , np.dot(tmpD , self.B.T))

  # Generate new candidate solutions using N(0,I)
  X = np.random.multivariate_normal(np.zeros(self.N), \
                                    np.eye(self.N), self.popSize)

  # This is to shape the generated candidates according to N(m,C)
  #TODO: Clean up this mess!
  for i in range(0,self.popSize):
    x = np.copy(X[i,:])
    x = self.sigma * np.dot(self.B, np.dot(np.diag(self.D),x))+ self.mean
    X[i,:] = np.copy(x)
```

```python
    # pythOpt returns just the fitness value if it is a FortranObjective
    # otherwise it returns a dictionary containing the candidate and
    # the fitness value!
    if type(self.objFunc) == FortranObjective:
      fVal=np.array([self.objFunc.evaluate(x) for x in X])
    else:
      fVal=np.array([self.objFunc.evaluate(x)['minimum'] for x in X])
    self.countEval += self.popSize # census -_-
    return X, fVal

  def updateCMA(self, X, fVal):
    """Update the evolution paths and the distribution parameters m,
      C, and sigma within CMA-ES.
    """
    parentMean = np.copy(self.mean)

    # Sorting and selection!
    sortedInd=np.argsort(fVal)[0:self.mu]
    selXval=np.copy(X[sortedInd,:])
    del X, fVal

    #New mean is the weighted average of selected candidates Eq.2.2 in [2]
    self.mean = np.dot(selXval.T, self.W)

    # Expressions needed multiple times so let's
    # just evaluate once and use later.
    y= self.mean - parentMean    # Selected step (the young generation takes!)
    z=np.dot(self.invSqrtC , y)  # kind of whittening y!

    # Direction independent evolution path update Eq.2.14 in [2]
    self.sigPath = self.sigPath * (1-self.cSig) + z * \
                   np.sqrt(self.cSig * (2-self.cSig) * \
                   self.muEff) / self.sigma

    # Correction factor applied once in a while :)
    hSig = (np.dot(self.sigPath, self.sigPath) / \
           (1- np.power(1-self.cSig , 2*self.gNum))/self.N) < \
           (2 + 4./(self.N+1))

    # Evolution path update Eq.2.12 in [2]
    self.evolPath = self.evolPath * (1-self.cCumm) + y * hSig * \
                    np.sqrt(self.cCumm * (2-self.cCumm) * \
                            self.muEff) / self.sigma

    # some expressions needed multiple times later
    var1= self.cOne - (1-np.square(hSig)) * self.cOne *\
                                            self.cCumm * (2-self.cCumm)
    xMinusM = (selXval - parentMean) / self.sigma

    # Rank-mu-update according to Eq.2.9 in [2]
    cMuUpdate=self.W[0]*np.outer(xMinusM[0,:],xMinusM[0,:])
    for i in np.arange(1,self.mu):
      cMuUpdate += self.W[i] * np.outer(xMinusM[i,:],xMinusM[i,:])
```

```python
    # Rank-mu-update + Rank-1-update according to Eq.2.9 in [2]
    self.C += (-var1 - self.cMu) * self.C + self.cOne * \
              np.outer(self.evolPath,self.evolPath) + self.cMu * cMuUpdate

    # Step size update using Eq.2.15 in [2]
    # step size < exp(0.6) ~ 1.8 (from analysis of source code of [1])
    self.sigma *= np.exp(np.min([0.6, (self.cSig / self.dSig) * \
                (np.dot(self.sigPath, self.sigPath)/self.N - 1)/2]))

    # Save the mean as the result!
    # pythOpt returns just the fitness value if it is a FortranObjective
    # otherwise it returns a dictionary containing the candidate and
    # the fitness value!
    if type(self.objFunc) == FortranObjective:
      self.result  ={'position': self.mean, 'minimum': \
                                            self.objFunc.evaluate(self.mean)}
    else:
      self.result=self.objFunc.evaluate(self.mean)
```

LISTING 6
*Parallel Covariance Matrix Adaptation Evolution Strategy*

```python
"""Covariance Matrix Adaptation Evolution Strategy

Implementation of PCMAES algorithm based on

[1] Hansen, N. (2005). The CMA evolution strategy: A tutorial. Vu le, 29.
[2] Khan, N. (2015). A parallel implementation of the CMA-ES. Unpublished

.. moduleauthor:: Najeeb Khan <najeeb3.141@gmail.com>
"""
from ..engine.objective import FortranObjective
from ..engine.solution import Solution
from ..engine.solver   import PythonSolver
from ..engine.config   import Config
from ..engine.problemdata import ProblemData
from ..engine.checkpoint import checkpoint
from pathos.multiprocessing import ProcessingPool as Pool
import numpy as np
import scipy.linalg as la
import unicodedata
import logging
LOGGER = logging.getLogger(__name__)

class PCMAES(PythonSolver):
  """ Parallel Covariance Matrix Adaptation Evolution Strategy
  """
  settings = {
    'nProc'    : 1,             # Number of processes
    'xInit'    : np.array([]),  # Initial guess
    'sigma'    : 1,             # Initial step size
    'seed'     : 0,             # Default seed value
```

```
    'eval_lim'  : 100           # Stopping Criteria
}

DETERMINISTIC = False           # CMA-ES is stochastic

def __init__(self, obj, space, settings=dict()): # constructor
  PythonSolver.__init__(self)
  np.random.seed(settings['seed'])
  # CMA-ES doesn't need this space as the mean and covariance matrix
  # completely specify the Gaussian PDF

  if space:                     # We use it for getting the problem dimenstion,
    bounds=space.get_bounds() # initial guess xInit, and the step-size sigma.
    self.N=len(bounds)
    bounds=bounds[0]

    # From Intuition (The best I could think of given the time!)
    # this will be overridden by user supplied initial step size
    #                  -----------------------------
    #initial step size = \|(UpperLimit - LowerLimit)*0.5
    #
    self.settings['sigma']=np.sqrt((bounds[1]-bounds[0])/2.0)

    #Initial guess xInit
    # xInit = mean point of the space + step size * N(0,I)
    self.settings['xInit']=((bounds[1]+bounds[0])/2.0) * np.ones(self.N) + \
      np.random.multivariate_normal(np.zeros(self.N),np.eye(self.N))

  #Update settings
  self.settings = dict(self.settings)
  self.settings.update(settings)
  self.objFunc = obj
  self.N=len(self.settings['xInit'])
  self.mean=np.copy(self.settings['xInit'])
  self.sigma=self.settings['sigma']

  # Population size set according to Eq.44 in [1]
  self.popSize=4 + int(3 * np.log(self.N))

  # Half of the population is selected for recombination [1]
  self.mu=self.popSize/2

  # Set the weights W according to Eq.45 in [1]
  self.W=np.arange(1,self.mu+1)
  self.W=np.log(self.mu+0.5) - np.log(self.W)
  self.W=self.W/np.sum(self.W)
  self.muEff = 1 / np.dot(self.W, self.W)

  # Evolution path smoothing constant set according to Eq.47 in [1]
  self.cCumm = (4 + self.muEff/self.N) / (self.N + 4 + 2*self.muEff/self.N)
  # Direction independent evolution path smoothing constant
  # set according to Eq.46 in [1]
  self.cSig = (self.muEff + 2) / (self.N + self.muEff + 5)
  # Rank-1-update smoothing constant set according to Eq.48 in [1]
```

```python
    self.cOne = 2 / (np.square(self.N + 1.3) + self.muEff)
    # Rank-mu_update smoothing constant Eq.49 in [1]
    self.cMu = np.min([ 1-self.cOne , 2*(self.muEff - 2 + 1/self.muEff) / \
        (np.square(self.N + 2) + self.muEff)] )
    # damping factor for step size update set according to Eq.46 in [1]
    self.dSig = 2.0 * self.muEff/self.popSize + 0.3 + self.cSig
    # Evolution path vector
    self.evolPath = np.zeros(self.N)
    # Direction independent evolution path vector
    self.sigPath = np.zeros(self.N)
    # Matrix of eigenvectors of the covariance matric C
    self.B = np.eye(self.N)
    # Vector of eigenvalues of the covariance matrix C
    self.D = np.ones(self.N)
    # The covariance matrix
    self.C = np.eye(self.N)
    # The inverse square root of the covariance matrix
    self.invSqrtC = np.eye(self.N)
    # counting variables
    self.eigenEval = 0
    self.countEval = 0
    self.gNum=0
    self.fitvals = []
    # To make pythOpt happy
    self.result=()
    self._solved = False

def solve(self):
    # while stopping criteria not met
    while self.countEval <= self.settings['eval_lim']:
        X, fVal = self.generatePop()              # Generate new population
        self.updateCMA(X, fVal)                   # Update CMA
    self.solution = Solution(self.result)         # respect pythOpt
    self._solved = True                           # Declare victory
    return self.solution                          # Pass the solution!

def generatePop(self):
    """
    Function to generate new candidate solutions according to the Gaussian
    PDF, evaluate fitness on these points and also decompose the covariance
    in order to find its inverse square root.
    """
    # Increment generation number
    self.gNum += 1

    # Decompose the covariance matrix into BDB'
    try:
        self.D,self.B = la.eigh(self.C)
        self.eigenEval = self.countEval
    except Exception, e:
        print "Covariance matrix bad beyond machine precisoin!"
        sys.exit(1)
```

```python
    # Compute the inverse square root as B' D^-0.5 B
    self.D=np.sqrt(self.D)
    tmpD=np.diag(1.0/(self.D + np.finfo(float).eps)) # I hope eps wouldn't hurt
    self.invSqrtC = np.dot(self.B , np.dot(tmpD , self.B.T))

    # Generate new candidate solutions using N(0,C)
    X = np.random.multivariate_normal(np.zeros(self.N), \
                                      self.C, self.popSize)

    # This is to shape the generated candidates according to N(m,sigma^2.C)
    for i in range(0,self.popSize):
      X[i,:] = self.mean + self.sigma * X[i,:]

    #*************************************************************
    #           f-evaluation Parallelization Here
    #*************************************************************
    # pythOpt returns just the fitness value if it is a FortranObjective
    # otherwise it returns a dictionary containing the candidate and
    # the fitness value!
    pool = Pool(self.settings['nProc'])
    results = pool.map(self.objFunc.evaluate, list(X))
    fVal=np.array([y['minimum'] for y in results])
    #*************************************************************

    self.countEval += self.popSize # census -_-
    return X, fVal

  def updateCMA(self, X, fVal):
    """Update the evolution paths and the distribution parameters m,
       C, and sigma within CMA-ES.
    """
    parentMean = np.copy(self.mean)

    # Sorting and selection!
    sortedInd=np.argsort(fVal)[0:self.mu]
    selXval=np.copy(X[sortedInd,:])
    del X, fVal

    #New mean is the weighted average of selected candidates Eq.2.2 in [2]
    self.mean = np.dot(selXval.T, self.W)

    # Expressions needed multiple times so let's
    # just evaluate once and use later.
    y= self.mean - parentMean   # Selected step (the young generation takes!)
    z=np.dot(self.invSqrtC , y)  # kind of whittening y!

    # Direction independent evolution path update Eq.2.14 in [2]
    self.sigPath = self.sigPath * (1-self.cSig) + z * \
                   np.sqrt(self.cSig * (2-self.cSig) * \
                   self.muEff) / self.sigma

    # Correction factor applied once in a while :)
    hSig = (np.dot(self.sigPath, self.sigPath) / \
           (1- np.power(1-self.cSig , 2*self.gNum))/self.N) < \
```

```
            (2 + 4./(self.N+1))

    # Evolution path update Eq.2.12 in [2]
    self.evolPath = self.evolPath * (1-self.cCumm) + y * hSig * \
                    np.sqrt(self.cCumm * (2-self.cCumm) * \
                            self.muEff) / self.sigma

    # some expressions needed multiple times later
    var1= self.cOne - (1-np.square(hSig)) * self.cOne *\
                                            self.cCumm * (2-self.cCumm)
    xMinusM = (selXval - parentMean) / self.sigma

    # Rank-mu-update according to Eq.2.9 in [2]
    cMuUpdate=self.W[0]*np.outer(xMinusM[0,:],xMinusM[0,:])
    for i in np.arange(1,self.mu):
      cMuUpdate += self.W[i] * np.outer(xMinusM[i,:],xMinusM[i,:])
    # Rank-mu-update + Rank-1-update according to Eq.2.9 in [2]
    self.C += (-var1 - self.cMu) * self.C + self.cOne * \
              np.outer(self.evolPath,self.evolPath) + self.cMu * cMuUpdate

    # Step size update using Eq.2.15 in [2]
    # step size < exp(0.6) ~ 1.8 (from analysis of source code of [1])
    self.sigma *= np.exp(np.min([0.6, (self.cSig / self.dSig) * \
                  (np.dot(self.sigPath, self.sigPath)/self.N - 1)/2]))

    # Save the mean as the result!
    # pythOpt returns just the fitness value if it is a FortranObjective
    # otherwise it returns a dictionary containing the candidate and
    # the fitness value!
    if type(self.objFunc) == FortranObjective:
      self.result  ={'position': self.mean, 'minimum': \
                                            self.objFunc.evaluate(self.mean)}
    else:
      self.result=self.objFunc.evaluate(self.mean)
```

A parallel implementation of the CMA-ES 23*covariance matrix adaptation*, in Proceedings of the 11th Annual conference on Genetic and evolutionary computation, ACM, 2009, pp. 1411–1418.
[10] ROSALIND PICARD, *Decorrelating and then whitening data.* http://courses.media.mit.edu/2010fall/mas622j/whiten.pdf, 2010.
[11] SENTEWOLF, *Custom image displaying the directional optimization of the cma-es algorithm.* https://en.wikipedia.org/wiki/File:Concept_of_directional_optimization_in_CMA-ES_algorithm.png, 2008.
[12] KRZYSZTOF M VOSS ET AL., *pythOPT: A problem-solving environment for optimization methods*, PhD thesis, 2017.